\documentclass[journal,twoside,web]{ieeecolor}
\usepackage{generic}
\usepackage{cite}
\usepackage{amsmath,amssymb,amsfonts}
\usepackage{graphicx}
\usepackage{hyperref}
\hypersetup{hidelinks=true}
\usepackage{textcomp}

\usepackage{subfigure}
\usepackage{caption}
\usepackage{float}
\usepackage{bm}
\usepackage{diagbox}
\usepackage{algorithm}
\usepackage{algorithmicx}
\usepackage{algpseudocode}
\usepackage{url}
\usepackage{indentfirst}
\usepackage{verbatim}
\usepackage{booktabs,multirow}
\usepackage{tikz}
\usetikzlibrary{spy}
\usepackage{cuted}
\usepackage{makecell}
\usepackage{bbding}
\usepackage{pifont}

\newcommand{\x}		{\mathbf{x}}

\newcommand{\E}     {\mathbf{E}}

\renewcommand{\P}	 {\mathbf{P}}
\newcommand{\omg}	 {\mathbf{\Omega}}

\newcommand{\X}		{\mathbf{X}}

\def\BibTeX{{\rm B\kern-.05em{\sc i\kern-.025em b}\kern-.08em
    T\kern-.1667em\lower.7ex\hbox{E}\kern-.125emX}}
\begin{document}
\title{Multi-scale Spatio-temporal Transformer-based Imbalanced Longitudinal Learning for Glaucoma Forecasting from Irregular Time Series Images}
\author{Xikai Yang, Jian Wu, Xi Wang, Yuchen Yuan, Ning Li Wang, and Pheng-Ann Heng, \IEEEmembership{Senior Member, IEEE}
\thanks{Corresponding authors are Xi Wang (xiwang@cse.cuhk.edu.hk) and Ning Li Wang (wningli@vip.163.com).}
\thanks{Xi Wang is with the Zhejiang Lab, Hangzhou, China, and also with the Department of Computer Science and Engineering, The Chinese University of Hong Kong, Hong Kong, China. (e-mail: xiwang@cse.cuhk.edu.hk)}
\thanks{Xikai Yang, Yuchen Yuan, and Pheng-Ann Heng are with the Department of Computer Science and Engineering, The Chinese University of Hong Kong, Hong Kong, China (e-mail: \{xkyang22, ycyuan22, pheng\}@cse.cuhk.edu.hk). }
\thanks{Jian Wu is with the Beijing Institute of Ophthalmology, Beijing Tongren Eye Center, Beijing Tongren Hospital, Capital Medical University, Beijing, China (e-mail: karena.wu@foxmail.com). }
\thanks{Ning Li Wang is with the Beijing Institute of Ophthalmology, Beijing Tongren Eye Center, Beijing Tongren Hospital, Capital Medical University, Beijing, China, and also with the Beijing Key Laboratory of Ophthalmology and Visual Sciences, Beijing, China (e-mail: wningli@vip.163.com). }
\thanks{Xikai Yang and Jian Wu contributed equally to this work.}
}

\maketitle

\begin{abstract}
	Glaucoma is one of the major eye diseases that leads to progressive optic nerve fiber damage and irreversible blindness, afflicting millions of individuals. Glaucoma forecast is a good solution to early screening and intervention of potential patients, which is helpful to prevent further deterioration of the disease. It leverages a series of historical fundus images of an eye and forecasts the likelihood of glaucoma occurrence in the future. However, the irregular sampling nature and the imbalanced class distribution are two challenges in the development of disease forecasting approaches.
	To this end, we introduce the Multi-scale Spatio-temporal Transformer Network (MST-former) based on the transformer architecture tailored for sequential image inputs, which can effectively learn representative semantic information from sequential images on both temporal and spatial dimensions. Specifically, we employ a multi-scale structure to extract features at various resolutions, which can largely exploit rich spatial information encoded in each image. Besides, we design a time distance matrix to scale time attention in a non-linear manner, which could effectively deal with the irregularly sampled data. 
	Furthermore, we introduce a temperature-controlled Balanced Softmax Cross-entropy loss to address the class imbalance issue.
	%address the imbalanced nature of the sequential fundus image dataset. 
	Extensive experiments on the Sequential fundus Images for Glaucoma Forecast (SIGF) dataset demonstrate the superiority of the proposed MST-former method, achieving an AUC of 98.6\% for glaucoma forecasting. Besides, our method shows excellent generalization capability on the Alzheimer’s Disease Neuroimaging Initiative (ADNI) MRI dataset, with an accuracy of 90.3\% for mild cognitive impairment and Alzheimer's disease prediction, outperforming the compared method by a large margin. A series of ablation studies further verify the contribution of our proposed components in addressing the irregular sampled and class imbalanced problems.
\end{abstract}


\section{Introduction}\label{sec:introduction}

\IEEEPARstart{G}{laucoma} is one of the most severe eye diseases which mainly causes progressive vision loss. Increasing intraocular pressure (IOP) results in optic nerve damage as well as the thickening of the retinal nerve fiber
layer (RNFL), which may block the information flow to the central brain~\cite{raghavendra2018deep} and lead to glaucoma. Nowadays, the rate of increase in the number of glaucoma patients is accelerating. According to the data provided by the World Health Organization (WHO), glaucoma has emerged as the second leading cause of blindness following cataracts~\cite{Quigley:06:tnp}. This situation is expected to become even more critical with the growth of the population and the aging trend.

Due to these adverse effects, researchers put considerable attention to the clinical diagnosis of glaucoma. In addition to employing visual field assessments to monitor the occurrence of glaucoma, early efforts for automated glaucoma detection relied on the observation and segmentation of the optic cup and optic disc regions within the eye for computing some diagnostic indicators~\cite{song2021deep,yin2011model,cheng2013superpixel,cheng2017quadratic}. However, such approaches demand a high standard of image data quality and necessitate further enhancements in terms of diagnosis accuracy.
Recently, along with the success of deep learning (DL) techniques in various areas, many works have focused on DL-based glaucoma prediction techniques. Through collecting extensive medical data, such as color fundus images, Optical coherence tomography (OCT), and visual field (VF) testing data, etc., researchers are capable of leveraging various prevailing deep learning and artificial intelligence frameworks to train models with promising performance. Song et al.~\cite{song2021deep} proposed the relation transformer with a union analysis of OCT and VF function for diagnosing glaucoma, which acquired a significant improvement over the existing single-model approaches. Compared to OCT imaging, fundus photography is a much faster and easier imaging technique with lower expense, which can be applied to a broader range of regions. There have been plenty of mature approaches that utilize fundus images for glaucoma diagnosis by far~\cite{Hemelings:20:apg,Singh:22:mdf}.

Regrettably, merely diagnosing patients based on their current imaging data falls short of clinical demands. A more meaningful objective is to predict glaucoma prior to its onset.
There is a deficiency of studies related to glaucoma forecasting. Anshul et al.~\cite{Anshul:20:pgb} collected around 66k fundus photographs and tried to predict glaucoma before its onset. An obvious drawback is that the lack of temporal data did not support modeling on time series images. Liu et al.~\cite{liu:20:deepGF} built the Sequential fundus Image for Glaucoma Forecast (\textbf{SIGF}) dataset, which contains $3671$ fundus images in total. These images were organized in chronological order, resulting in $405$ image sequences with different sequence lengths (from $6$ to $28$). The authors design an LSTM-based network to capture the time-variant feature and incorporate the time interval information. The method takes fundus image sequences containing five consecutive visits as input, and generates the probability of glaucoma occurrence for the 6th image.
Recently, Hu et al.~\cite{xiaoyan:23:glim} proposed the GLIM-Net to improve the predictive accuracy of the SIGF dataset by adopting the encoder-decoder structure of the classical transformer. Both two methods employed active convergence (AC) to handle the class imbalanced problem.
%Experimental results on the SIGF dataset demonstrate the superiority of their proposed approach over previous methods. 
Nonetheless, the aforementioned two works for glaucoma forecasting primarily emphasize modeling temporal information while overlooking the extraction of representative semantic features from informative regions encoded in each individual fundus image. 
For instance, GLIM-Net first uses an embedding module to transform the whole image into a token vector, which is then fed into subsequent transformer modules to model the temporal relation. Without identifying the discriminative area within each image, it might be difficult for the model to build a strong correlation between the adjacent images, and thus fails to capture the subtle progression of the disease. Besides, the AC strategy involves sophisticated configurations of a number of hyperparameters, which might impede the application and popularization of these methods in clinical practice.

Therefore, in this work, we present an innovative imbalanced longitudinal learning method, \textbf{M}ulti-scale \textbf{S}patio-\textbf{T}emporal trans\textbf{former} (MST-former) framework for glaucoma forecasting. It splits the time series images into fixed-length patches rather than converting the whole image into a single vector. We introduce a space-time positional encoding module tailored for spatio-temporal representation, which exploits the intra- and inter-image position information. 
To enhance the discrimination ability of the model, we use the multi-scale technique from the classical computer vision realm to extract spatial information across different scales. Besides, we impose a time-aware scaling on the temporal attention to deal with the irregular sampling issue, and propose temperature-controlled Balanced Softmax Cross-entropy loss to alleviate the severe class imbalanced problem. We conduct extensive experiments on two medical datasets for disease forecasting in comparison with several state-of-the-art methods, and also perform ablation study to investigate the efficacy of the proposed components and rationality of the hyperparameter settings.
%Grounded in the transformer architecture, our approach models both the temporal and spatial dimensions within each transformer block. Moreover, it employs time-aware attention to compute attention weights along the temporal dimension in a non-linear manner. 
%We further introduce a positional encoding module tailored for spatio-temporal representation, encompassing components for both temporal and spatial positional encoding. More specifically, we integrate multi-scale techniques from the classical computer vision realm to model and extract spatial infor- mation across different scales. We also propose a Balanced Softmax Cross-entropy loss with temperature control to deal with the imbalanced situation of the SIGF training dataset. The experimental results on the SIGF dataset demonstrate that our proposed method outperforms other compared approaches by a large margin. We also validate the effectiveness of our proposed method on the Alzheimer’s Disease Neuroimaging Initiative (ADNI) dataset. The results show that the MST-former exhibits strong performance in early forecasting from longitudinal brain MRI data
We summarize our contributions to this work as follows:
\begin{itemize}
	\item We propose the multi-scale spatio-temporal transformer network (MST-former) to model the spatial and temporal information simultaneously using multi-scale encoder-decoder blocks, aiming at forecasting glaucoma from irregular time series fundus images.
	\item We design time-aware temporal attention and multi-scale structure to reinforce the representation learning along the time and space dimensions, respectively, which can effectively address the irregular sampling issue.
	\item We improve the Balanced Softmax Cross-entropy loss with temperature control to handle the class imbalanced issue in the training set. Compared with previous works that employ multi-stage training with the AC strategy, the $\tau$-control Balanced Softmax Cross-entropy loss enables end-to-end training with the whole training samples, which is more efficient and elegant.
	\item We extensively evaluate the proposed method on the SIGF dataset. The experimental results show that the proposed MST-former achieves state-of-the-art performance in glaucoma forecasting, outperforming other methods significantly. The validation experiments on the Alzheimer's Disease Neuroimaging Initiative (ADNI) dataset further confirm the strong robustness and generalization capability of our method.
\end{itemize}

\section{Related Works}\label{sec:relate_work}

In this section, we will elaborate on the previous studies on disease forecasting and the most relevant techniques to this work as follows.
\subsection{Early Disease Forecast}
The early prediction of disease occurrence and the trend of progression hold paramount significance in clinical practice. Recently, with the advancements in deep learning, there has been a proliferation of research aiming at utilizing retrospective clinical data to forecast future states or the probability of disease onset.

Toma{\v{s}}ev et al.~\cite{tomavsev2019clinically} developed a deep-learning method for predicting future acute kidney injury. They built a deep recurrent highway network embedded with Recurrent Neural Network (RNN) modules. This method was evaluated on the collected longitudinal electronic health records (EHR) dataset, and proven effective in assessing the risk of disease onset. Rather than EHR, some studies leveraged historical sequential imaging data to forecast current  status~\cite{santeramo2018longitudinal,tao2022longitudinal,gao2020time}. 
Santeramo et al.~\cite{santeramo2018longitudinal} applied the time-modulated Long-Short-Term-Memory (LSTM) networks to model a sequence of chest X-ray images and proved the effectiveness of recognizing the chest X-ray. For instance, Tao et al.~\cite{tao2022longitudinal} employed the LSTM-based classifier and predicted the invasiveness of lung nodules based on retrospective CT data. To address the issue of irregular sampling in temporal image data, Gao et al.~\cite{gao2020time} modified the LSTM model and introduced the Distanced LSTM approach, which incorporates time-distanced gates to enable explicit modeling of sampling time intervals. They conducted experiments on the lung CT dataset and achieved exceptional results. 
Note that the aforementioned RNN-based approaches mainly focus more on the latest data, and the performance is highly related to the quality of the feature embedding provided by the feature extraction module.
Lin et al.~\cite{lin2022multi} proposed the MMSNet to predict future Primary open-angle glaucoma (POAG), which follows the general diagnosis workflow in routine clinical practice. However, MMSNet only focused on the similarity between the baseline and follow-up images and ignored the importance of time series information. 

Given the widespread successful application of transformers in modeling time sequences such as language, researchers are inspired to apply transformers to deal with sequential medical images~\cite{sarasua2021transformesh,li2023time,li2023longitudinal,xiaoyan:23:glim}. Li et al.~\cite{li2023time} revised the structure of the vision transformer (ViT) and proposed the time-aware ViT, which enables scaling self-attention with the time-distance matrix. Similarly, Hu et al.~\cite{xiaoyan:23:glim} employs the time-sensitive multi-head self-attention in the conventional transformer and presents the GLIM-Net for glaucoma forecast. In these works, each of sequential image should be translated into an embedding vector as the input tokens for the transformer. A prominent shortcoming of these methods is that the model performance is heavily dependent on the feature embedding module. Without explicitly exploiting the spatial information of images, the model might fail to capture the discriminative features (e.g., the subtle symptoms during early progression of disease) from individual images.

\subsection{Transformer with Spatial-temporal Representation}
Despite the significant success of transformers in natural language processing, how to effectively integrate the characteristics of transformers into various computer vision tasks still remains a substantial challenge. %One of the problems is how to adeptly leverage transformers to extract spatio-temporal representations. 
Vision Transformer (ViT)~\cite{dosovitskiy2020image} is one good example of applying the transformer to image processing. However, when it comes to extending ViT to handle inputs with temporal sequential formats, such as videos or other time-dependent image sequences, some modifications are necessary. One of the most straightforward ways is to convert each image into a token through an embedding module in advance so that the transformer gets a series of regular tokens as input~\cite{sarasua2021transformesh,arnab2021vivit}. Other works focus on how to effectively fuse the temporal and spatial dimensions within the transformer framework. For example, Bertasius et al.~\cite{bertasius2021space} discussed several mutations of the conventional self-attention computation and presented a convolution-free approach by applying temporal attention and spatial attention separately within each block. Ryoo et al.~\cite{ryoo2021tokenlearner} introduced an adaptively learned token for video understanding tasks, which can automatically exploit important tokens from the input. Feichtenhofer et al.~\cite{feichtenhofer2022masked} extended the Mask Autoencoders (MAE)~\cite{he2022masked} to spatio-temporal representation learning area, which can outperform some supervised methods. A recent work called UniFormer~\cite{li2201uniformer} combined the convolution neural networks (CNNs) and Vision Transformer in a way that captures long-range dependency via self-attention while minimizing the impact of redundancy among video frames. 
However, most of these works lack modules for modeling temporal information. Especially when facing irregularly sampled sequential images as inputs, it is obviously irrational to assign equal weights to each frame image. Hence, we introduce a novel framework, dubbed MST-former, to explicitly represent sequential images in both spatial and temporal dimensions. Additionally, the MST-former incorporates multi-scale techniques to enhance spatial feature extraction, while the time-aware temporal attention module is designed to fit irregular sequential image inputs.

\section{Methodology}\label{sec:method}

\begin{figure}[htbp]
	\centering
	\includegraphics[width=0.5\textwidth]{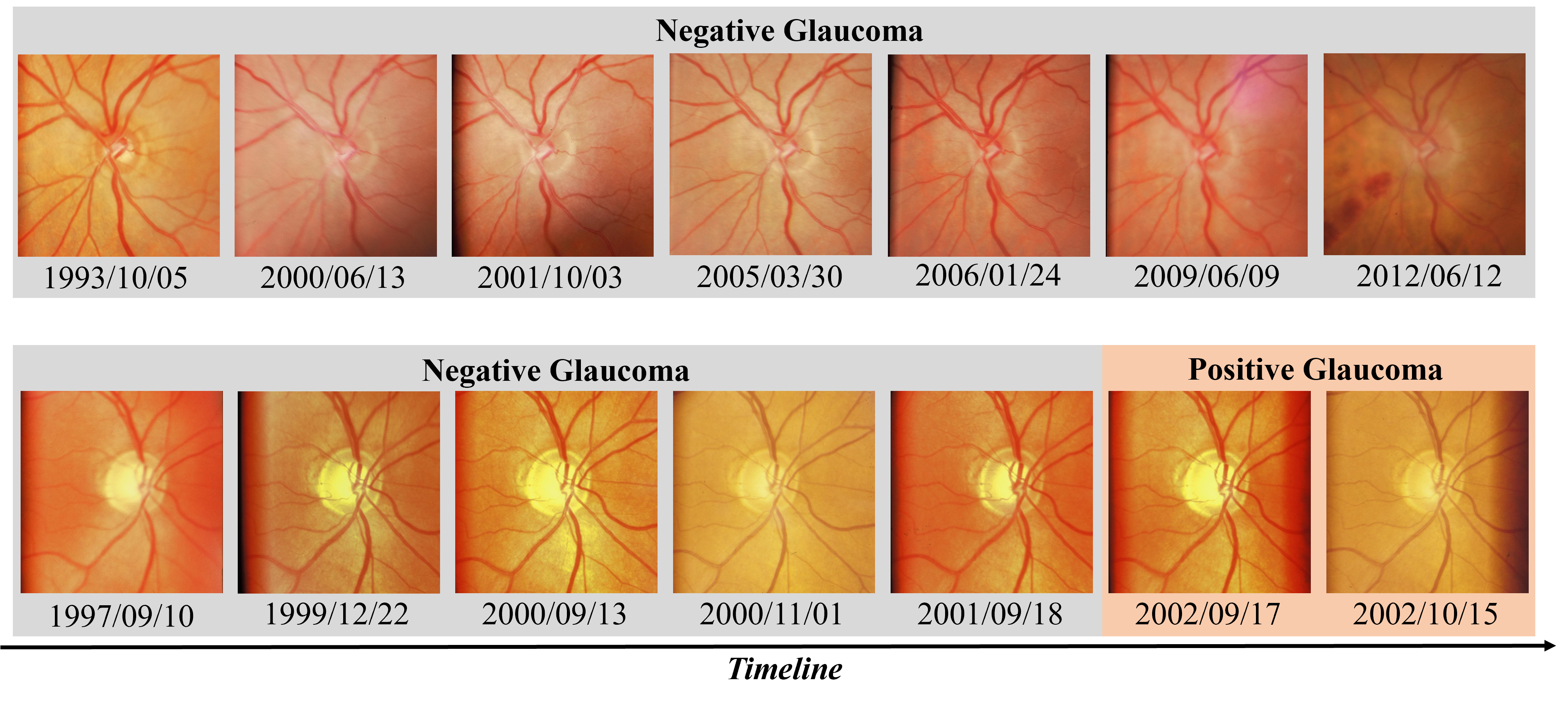}	
	\caption{Examples of the sequential fundus images in the SIGF database. The upper panel shows a time-invariant sequence, showing that the patient's eye keeps the negative status across all time points. The lower panel is a time-variant sequence, showing that the patient's eye converts from normal to glaucoma at \textit{2002/09/17}.}
	\label{fig:seq_fundus_case}  
\end{figure}

In this section, we first give a mathematical description of the glaucoma forecasting task followed by the introduction of our proposed framework in detail. Fig.~\ref{fig:seq_fundus_case} shows one case of sequential fundus images in the SIGF database. Specifically, we represent a series of fundus images (e.g., L images) collected from the same patient's eye as $\{\X_l, 1\leq l \leq L\}$, and $\X_l \in \mathbb{R}^{224\times 224 \times 3}$ denotes the image scanned at the $l$th time point. Each scan is associated with one label $Y_l \in \{0,1\}$, where $0$ and $1$ stand for the normal and glaucomatous case, respectively. Our goal is to output the probability of positive glaucoma of the next image (i.e., $\X_{L+1}$). In other words, we formulate the glaucoma forecasting problem as $\tilde{Y}_{L+1} = \mathcal{F}(\X_{L},\X_{L-1}\cdots, \X_{1}, Y_{L}, Y_{L-1},\cdots, Y_{1})$, where $\mathcal{F}$ is our proposed MST-former framework.
% and $K$ denote the length of the observation window. 

In this work, we propose the multi-scale spatio-temporal transformer network (MST-former) to predict the status of patients at the next time point. The MST-former model enables handling input being sequential images and learns patch representations along the space and time dimensions simultaneously.
We also ameliorate the Balanced Softmax Cross-entropy loss~\cite{ren2020balanced} function to conquer the challenges brought by the imbalanced training set. The detailed description of our proposed method is presented in the following subsections.
\subsection{Vanilla Transformer} \label{sec:naive_transformer}
\begin{figure}[htbp]
	% \begin{strip}
		\centering
		\includegraphics[width=0.3\textwidth]{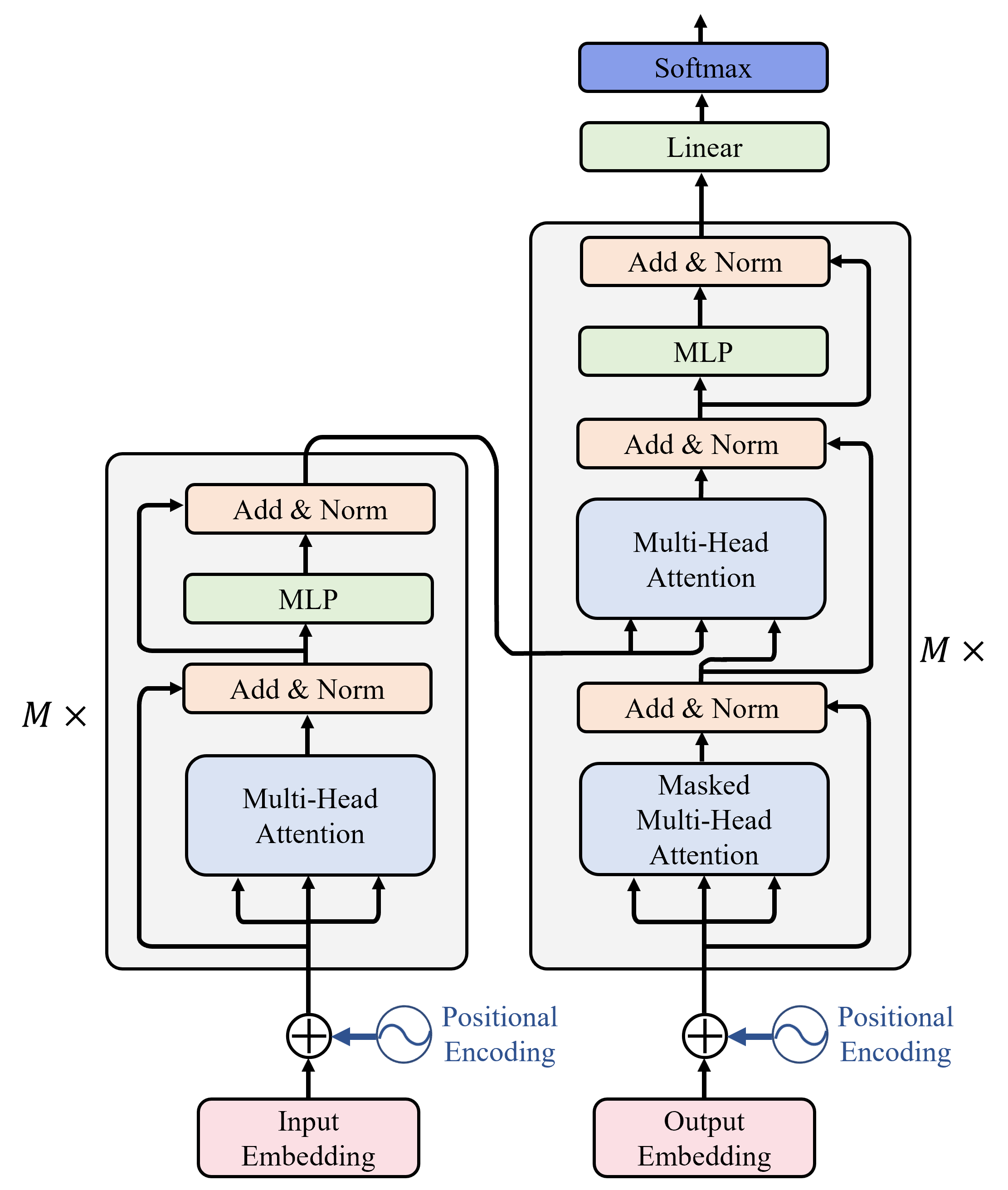}	
		\caption{Architecture of the conventional transformer.}
		\label{fig:transformer_structure}  
		% \end{strip}
\end{figure}

As shown in Fig.~\ref{fig:transformer_structure}, the conventional transformer~\cite{vaswani2017attention} is composed of a stack of $N$ encoder blocks and $N$ decoder blocks. Each image is split into a sequence of fixed-size patches, followed by linear embedding. Then the position encoding is imposed to make use of the order information of patch vectors before being fed into the standard transformer encoder.
In each encoder block, the input is expanded into query, key, and value components, which are operated by the multi-head self-attention (MHSA) pipeline, followed by a fully connected feed-forward (FF) module with residual brunch. 
During the decoding stage, each decoder block comprises the identical MHSA and FF modules as the encoder block. In addition, a masked self-attention module is introduced to maintain causality of the sequence, ensuring that the prediction at the current position depends only on the known outputs at positions before.

\subsection{MST-former Framework} \label{sec:general_struc}
Unlike previous transformer-based studies that take the temporal information as positional encoding embedding and merely focus on imposing attention weights on the sequential feature vectors~\cite{xiaoyan:23:glim,li2023time}, we employ the spatio-temporal attention technique~\cite{bertasius2021space} to fully exploit the rich information locally and globally encoded in the sequential data. On the one hand, spatial attention can identify the informative regions within each individual image. On the other hand, temporal attention can help capture the disease progression cue from the longitudinal data.

\begin{figure*}[htbp]
	% \begin{strip}
		\centering
		\includegraphics[width=0.8\textwidth]{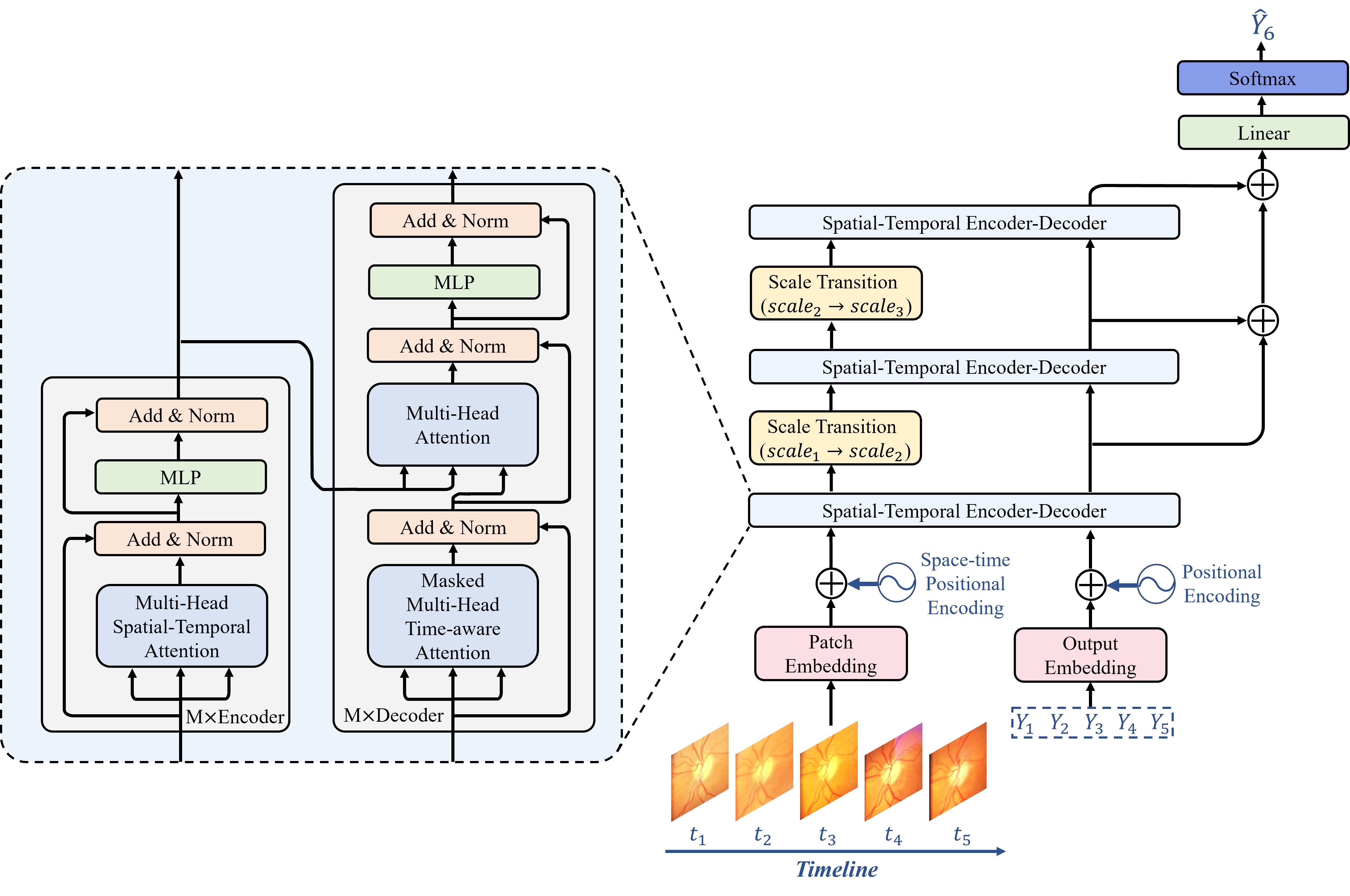}	
		\caption{Illustration of the proposed multi-scale spatio-temporal transformer network (MST-former), which includes $3$ scales. Within each scale, there are $N$ encoder and decoder blocks. The input of the encoder comprises the patch embedding and the space-time positional encoding information, while the input of the decoder contains the output embedding together with its positional embedding. The circle plus symbol represents the element-wise addition.}
		\label{fig:MST-former_structure}  
		% \end{strip}
\end{figure*}

Fig.~\ref{fig:MST-former_structure} shows the structure of the MST-former network, which contains stacked spatial-temporal encoder-decoder blocks of $S$ scales. It takes a sequence of images as input and output the probability of the next visit. 
More specifically, in the encoder-decoder structure of the first scale, the patch embeddings (i.e., \{$x_{i,n}$\} where $i\in [1,L]$ and $n\in [1,N]$, $N$ is the number of patches within one image) are fed into the multi-head spatial-temporal attention (MHSTA) module to proceed self-attention from both the spatial and temporal dimensions, resulting \{$\overline{x}_{i,n}^{1}$\}. In parallel, the corresponding labeling embedding (i.e., \{$y_{i}$\} where $i\in [1,L]$) is processed by the masked multi-head time-aware temporal self-attention module in the decoder, resulting \{$\hat{y}_{i}^{1}$\}. Next, \{$\overline{x}_{i,n}^{1}$\} in conjunction with \{$\hat{y}_{i}^{1}$\} is used to execute self-attention. The resulting weighted features is then passed into the subsequent decoder layers to yield the output, \{$\overline{y}_{i}^{1}$\}. After scale transition on the encoder's output \{$\overline{x}_{i,n}^{1}$\}, the condensed feature \{$\tilde{x}_{i,n}^{1}$\} and the decoder's output \{$\overline{y}_{i}^{1}$\} are fed as input to the encoder-decoder structure at scale 2. These steps are repeated $S-1$ times. 

The outputs of the decoder module at different scales (i.e., $\overline{y}^{s}$, $s\in [1,S]$ ) are aggregated and eventually input into the classification head to predict the state at the next time stage.

Distinct from the vanilla transformer, our improvements concentrate on the following aspects: a) We design a novel space-time positional encoding module to encode the intra-image and inter-image positional information of the input patches.
%To achieve the transition from 2D image-level positional encoding to 3D space-time-level positional encoding, we leverage a space-time positional encoding module to represent the positions of embedding patches. 
b) We operate time-aware scaling to the temporal attention in the MHSTA module of the encoder and the masked MHSA of the decoder.
%We use multi-head spatial-temporal attention to replace the ordinal self-attention module in the encoder stage. 
c) We construct a hierarchical structure of multiple scales to capture representative features at different resolutions. 
%d) The time-aware attention mechanism is applied to all attention-generation processes along the temporal dimension. 
Next, we provide a detailed description of these components in the following subsections.

\subsubsection{Patch Embedding and Space-time Positional Encoding}
For each image $\X_l \in \mathbb{R}^{H\times W \times C}$ within the image sequence, we transform it into a sequence of flattened 2D patches following the vision transformer~\cite{dosovitskiy2020image}. Specifically, we convert each patch into a $d_m$ dimensions vector by a convolution layer with the kernel size of $p\times p$, stride of $p$, and output channel of $d_m$:
\begin{equation}\label{eq:patch_embed}
	\begin{aligned}
		\{\x_{l,n}\}=\P(\X_l), \;\; 1\leq l \leq L,\; 1 \leq n \leq N.
	\end{aligned}
\end{equation}
where $\P$ denotes the patch embedding operation, $\x_{l,n} \in \mathbb{R}^{d_m}$ means the embedding of the $n$th patch of the $l$th image, and $N=\frac{HW}{p^2}$ refers to the total number of patches of each image.

Due to the permutation-invariant property of self-attention~\cite{wang2022lilt}, how to leverage the order of the tokens as part of input information is one challenging task across all transformer-based model designs. In this work, we propose the space-time positional encoding (STP) which is compatible with the three-dimensional token input. The mathematical representation is given as in Eq.~\eqref{eq:time_space_pos}, 
\begin{equation}\label{eq:time_space_pos}
	\left \{
	\begin{aligned}
		\text{STP}(l,n)[2i] &= \sin(\frac{\Delta t_{l,1}}{10000^{\frac{2i}{d_m}}}) + \sin(\frac{n}{10000^{\frac{2i}{d_m}}})\\
		\text{STP}(l,n)[2i+1] &=  \cos(\frac{\Delta t_{l,1}}{10000^{\frac{2i}{d_m}}}) + \cos(\frac{n}{10000^{\frac{2i}{d_m}}})
	\end{aligned}
	\right.
\end{equation}
where $\Delta t_{l,1} = |t_l - t_1|$ that indicates the time interval at the $l$-th visit ever since the first examination. $d_m$ represents the size of embedding vectors, $i$ ranges from $0$ to $d_m/2-1$ (i.e., $i\in \{0,1,...,d_m/2-1\}$), and $n$ is the patch index. As shown in Fig.~\ref{fig:MST-former_structure}, in the encoder, we input the sequential patch embeddings along with their corresponding space-time positional encoding information, whereas in the decoder, we impose the time positional encoding information on the output embedding~\cite{xiaoyan:23:glim}.

\subsubsection{Multi-head Spatial-Temporal Attention}
\begin{figure}[htbp]
	% \begin{strip}
		\centering
		\includegraphics[width=0.25\textwidth]{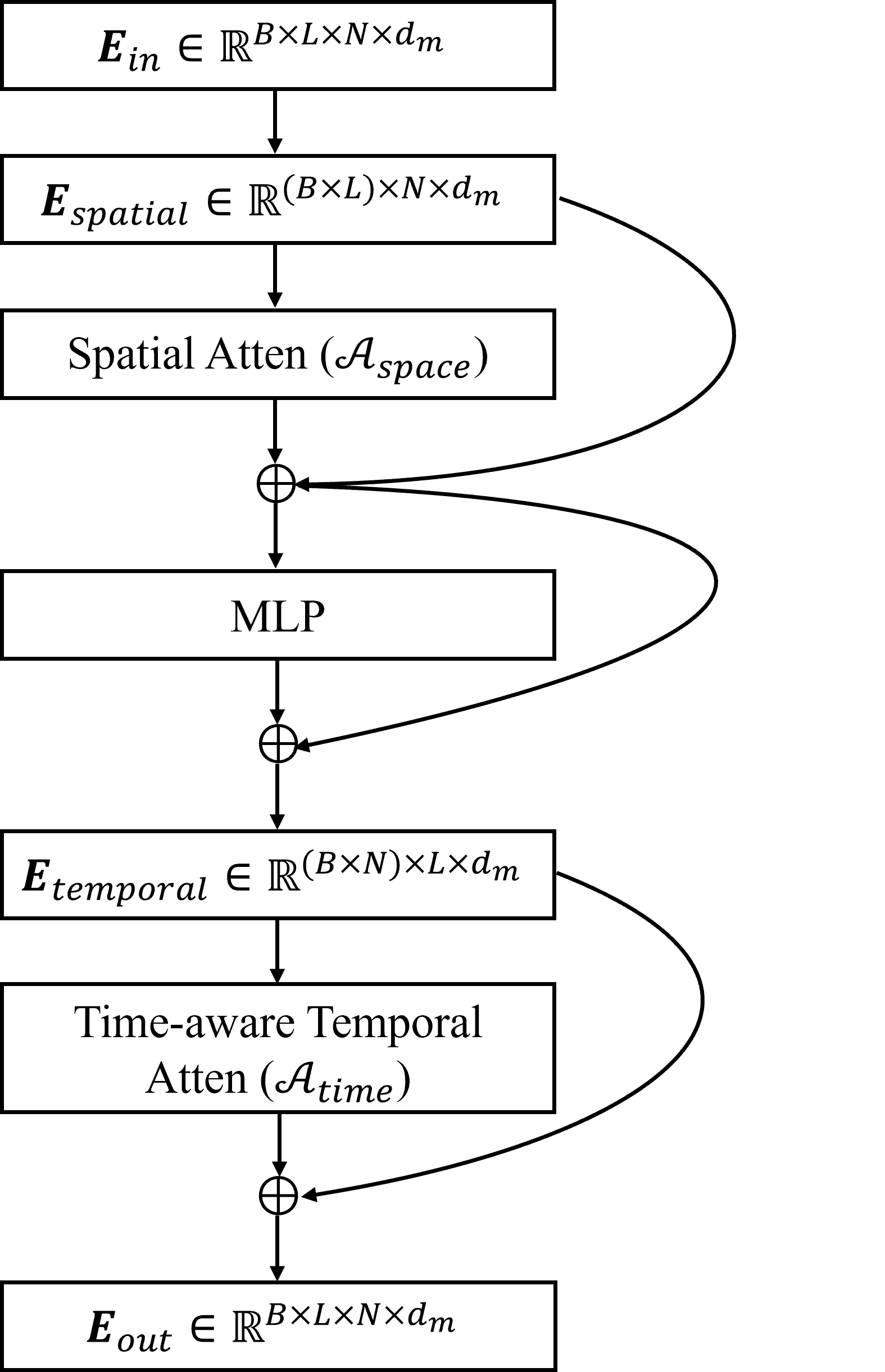}	
		\caption{Spatial-temporal self-attention block in MST-former. Two-level self-attention steps (spatial self-attention and time-aware temporal self-attention) are included to calculate attention scores along the space and time dimensions.}
		\label{fig:space_time_atten}  
		% \end{strip}
\end{figure}

%\blue{After blending the input sequential images with space-time positional information, we need to further extract their spatio-temporal features during the encoding stage.} 
Bonded with space-time positional information, the sequential data is fed into the encoder for feature extraction.
Here, we propose the multi-head spatial-temporal attention block to compute the attention scores along the space and time dimensions sequentially. 
The input of the multi-head spatial-temporal attention block is denoted as $\E \in \mathbb{R}^{B \times L \times N \times d_m}$, where $B$ denotes the batch size, $L$ stands for the sequential length, and $N$ is the number of patches of one image. 
Fig.~\ref{fig:space_time_atten} shows the spatial-temporal self-attention block used in MST-former. Specifically, inside the proposed block, we reshape the input $\E$ by swapping the space and time dimensions alternatively, following calculating corresponding attention scores. We initially merge the batch dimension and the time dimension, obtaining $\E_{spatial} \in \mathbb{R}^{(B\times L) \times N\times d_m}$ as the input for the spatial attention calculation step.
The spatial attention is operated following the standard self-attention operation, which is defined as in Eq.~\eqref{eq:atten}.
\begin{equation}\label{eq:atten}
	\begin{aligned}
		&\mathcal{A}_{space}(\mathbf{Q}, \mathbf{K}, \mathbf{V}) = {\rm softmax}(\frac{\mathbf{Q}\mathbf{K}^T}{\sqrt{d_m}})\mathbf{V}, \\
		\mathbf{Q}=&\mathcal{F}^Q(\E_{spatial}), \; \mathbf{K}=\mathcal{F}^K(\E_{spatial}), \;\mathbf{V}=\mathcal{F}^V(\E_{spatial}).
	\end{aligned}
\end{equation}
$\mathcal{F}^Q$, $\mathcal{F}^K$, $\mathcal{F}^V$ are three linear functions that convert input matrix into query matrix ($\mathbf{Q}$), key matrix ($\mathbf{K}$), and value matrix ($\mathbf{V}$), respectively.

% \subsubsection{Time-aware Scaling Attention}
After finishing the spatial self-attention, we reshape the embedding matrix as $\E_{temporal} \in \mathbb{R}^{(B\times N)\times L \times d_m}$ and perform the time-aware temporal self-attention along the time direction. 

Note that the time series observations usually have non-uniform time intervals between successive observations, which is a common phenomenon in healthcare. How to align features of the observations with distinct sampling ratios is quite important for developing robust and reliable models.
For disease progression, considering that recent events usually have a greater impact on the present than distant events, it is reasonable to assign different weights to the computed attention at different time points for feature alignment. 
Similar to~\cite{li2023time}, we leverage the time distance $\Delta t_{i,j}$ to scale the attention scores.
The scaling value between any two visits $\X_i$ and $\X_j$ is calculated by
\begin{equation} \label{eq:t_matrix}
	\begin{aligned}
		&\quad \omg_{i,j} = \frac{1}{1+e^{\alpha\Delta t_{i,j}-\beta}} 
		\\
		&\mathrm{s.t.} \quad 1\leq i \leq L; 1\leq j \leq i
	\end{aligned}
\end{equation}
%Eq.~\eqref{eq:t_matrix} gives the formula to obtain the scaling value between $\X_i$ and $\X_j$. 
where $\alpha$ and $\beta$ are hyper-parameters that control the extent of time scaling.
Hence, the time-aware temporal self-attention is as follows:
\begin{equation} \label{eq:tsa}
	\begin{aligned}
		&\mathcal{A}_{time}(\mathbf{Q}, \mathbf{K}, \mathbf{V}, \mathbf{\omg}) = {\rm softmax}(\frac{(\mathbf{Q}\mathbf{K}^T)*\omg}{\sqrt{d_m}})\mathbf{V},
		\\
		&\mathbf{Q}=\mathcal{F}^Q(\E_{temporal}), \; \mathbf{K}=\mathcal{F}^K(\E_{temporal}, \;\mathbf{V}=\mathcal{F}^V(\E_{temporal}),
	\end{aligned}
\end{equation}
where ``$*$" denotes the element-wise multiplication operation between two matrices.
The multi-head self-attention is identical to the conventional transformer. The input queries, keys, and values $\{\mathbf{Q}, \mathbf{K}, \mathbf{V}\}\in \mathbb{R}^{B \times L \times N \times d_m}$ are divided into $Z$ heads $\{\mathbf{Q}_i, \mathbf{K}_i, \mathbf{V}_i, 1\leq i \leq Z\} \in \mathbb{R}^{B \times L \times N \times d_m/Z}$. The spatial-temporal attention block works on them in parallel and the final outputs from all heads are stacked together with the same shape as the input. 

\subsubsection{Multi-scale Encoder-decoder Architecture}
\begin{figure*}[htbp]
	% \begin{strip}
		\centering
		\includegraphics[width=0.7\textwidth]{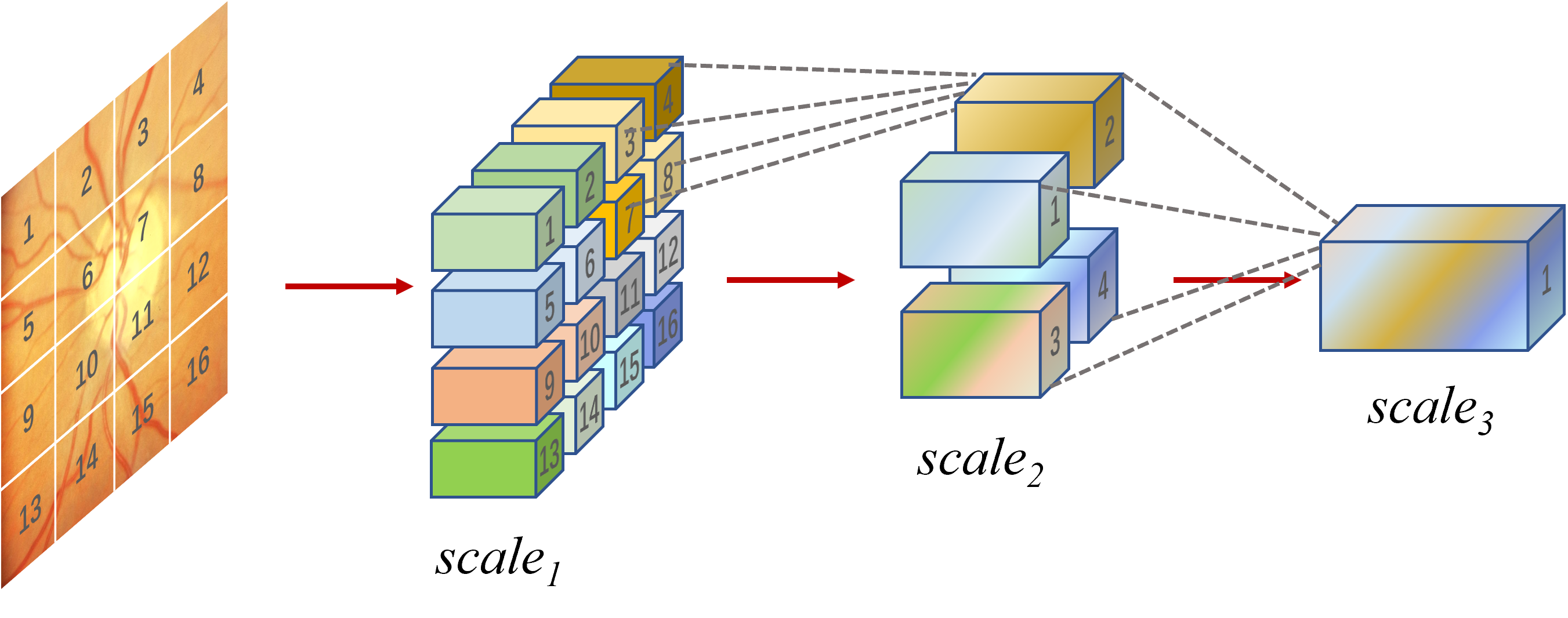}	
		\caption{Sample illustration of the Multi-scale structure of MST-former. Here, we present $3$ scales. During the process of scale transition, tokens that are topologically adjacent in a $2$ by $2$ format are merged together. Note that this multi-scale diagram depicts only one single image. For the input to be sequential medical images, it is necessary to perform the same operations in parallel for each individual image within the sequence.}
		\label{fig:MST_multiscale}  
		% \end{strip}
\end{figure*}

Unlike the conventional transformer keeping the same resolution across all encoder attention blocks, we employ multi-scale architecture to reinforce the representation ability in image space and the resolution can be shrunken hierarchically. Fig.~\ref{fig:MST_multiscale} gives one example related to our multi-scale structure applied to the encoder part. During the stage of scale transition, we reshape the $\E_{out}$ into the shape of $B\times L \times \frac{H}{p} \times \frac{W}{p} \times d_m$ and for each feature sub-matrix of size $\frac{H}{p} \times \frac{W}{p} \times d_m$, we perform patch embedding operations with a patch size of $\gamma$, so that $\gamma \times \gamma$ topologically adjacent tokens are merged together, which largely lowers the spatial resolution of the underlying visual data while enabling the network to learn from local to global features~\cite{fan2021multiscale}.

As shown in Fig.~\ref{fig:MST-former_structure}, at the same scale, the input of the decoder part 
% ($\D^i \in \mathbb{R}^{B \times L \times d_m}$) 
goes through a masked multi-head time-aware temporal self-attention module with the same computational formula as Eq.~\eqref{eq:tsa}. Subsequently, the resulting output is combined with the reshaped encoder's output to calculate attention and produce predictions for this scale, which can be denoted as  $\overline{y}^{s}$. Finally, we aggregate the outputs from all scales as $\overline{y} = \sum_{s=1}^{S} \overline{y}^{s}$ and we utilize a classifier head to obtain the ultimate model prediction probabilities.

\subsection{Balanced Softmax Cross-entropy Loss with Temperature Control}
Note that a severe class imbalance issue exists in the glaucoma forecasting dataset (i.e., the ratio of positive and negative samples is around 19), which will cause model bias and negatively impact its performance.
To solve this problem, the previous works~\cite{liu:20:deepGF,xiaoyan:23:glim} mainly adopt the active convergence (AC) strategy to dynamically drop training samples with a lower loss value during the training process. However, this strategy has two limitations. First, discarding partial training samples will break the original training scheme and easily make the training process unstable. Second, it is difficult to determine when to take the AC strategy and how many samples to be discarded. In~\cite{liu:20:deepGF,xiaoyan:23:glim}, there are a number of hyper-parameters involved, which are determined in an empirical manner.
%they select these parameters in an empirical manner, which involves a number of hyper-parameters. 
For example, GLIM-Net has to manually tune not only the number of sample dropout iterations, but also the learning rate and the number of discarded samples in each sample dropout process. From the experimental results reported in ~\cite{xiaoyan:23:glim}, the model performance is highly sensitive to the value of the hyper-parameters. 

Recently, the Balanced Softmax Cross-entropy loss was proposed in~\cite{ren2020balanced} to address the long-tailed problem, and was verified more effective than many other end-to-end training methods (e.g., class-balanced weighting, class-balanced sampling, Focal Loss~\cite{lin2017focal}, class-balanced loss~\cite{cui2019class}, LDAM loss~\cite{cao2019learning}, Equalization Loss~\cite{tan2020equalization}) and decoupled training methods (e.g., cRT~\cite{kang2019decoupling} and LWS~\cite{kang2019decoupling}) in long-tailed tasks with moderate imbalance ratios. However, for datasets with an extremely large imbalance, the optimization process becomes difficult~\cite{ren2020balanced}. To this end, we introduce a temperature scalar $\tau$ in the Balanced Softmax Cross-entropy to control the extent of class weights, which particularly focuses on the severe class imbalance issue. The improved Balanced Softmax Cross-entropy loss is as follows:
\begin{equation}\label{eq:balancedCE}
	\mathcal{L} = - \sum_{i=1}^k y_i \log \bigg( \frac{n_{i}^{\tau}e^{\tilde{y}_i}}{\sum_{j=1}^k n_j^{\tau} e^{\tilde{y}_j}}  \bigg)
\end{equation}
where $y_i$ and $\tilde{y}_i$ denote the ground truth label and network output for the $i$th class, respectively. $n_i$ represents the number of samples attributed to the $i$th class, which is essential to mitigate label distribution shifts between the training and test sets. 
As $\tau$ increases, the influence of the distribution in the training dataset on the loss function will gradually become more significant.

\section{Experiments}\label{sec:experiment}

\subsection{Dataset and Evaluation Metrics}
In this study, we evaluate the performance of our proposed MST-former framework on two sequential medical datasets, one is the SIGF dataset, and the other is the ADNI dataset. 
\subsubsection{SIGF dataset}
The SIGF dataset is originally revealed in~\cite{liu:20:deepGF} and is the first longitudinal fundus dataset constructed for glaucoma forecasting given a series of inspection imaging data.
It is comprised of retrospective color fundus pictures from $405$ eyes, i.e., $405$ image sequences or $3671$ fundus images in total.
The time series images is not evenly sampled across all patients (e.g., those in Fig.~\ref{fig:seq_fundus_case}).
The average length of all sequences is around $9$.
We classify the image sequences into two types, time-invariant sequences which contain the transition from negative status to positive status, and time-variant sequences that keep the negative label at all times. The overall numbers of time-variant and time-invariant sequences inside the SIGF dataset are $37$ and $368$, respectively. Table~\ref{table:SIGF_stat} presents the detailed statistics summary of the SIGF dataset.
We keep the same split as in~\cite{liu:20:deepGF,xiaoyan:23:glim}, i.e., the dataset is divided into three sets for training ($300$), validation ($35$), and testing ($70$). Specifically, the average time span length of the training set, the validation set and the testing set are $13.1$, $16.0$ and $13.8$ years, respectively.
\begin{table}[!h]
	\centering
	\caption{Summary statistics of the SIGF dataset before and after data augmentation} 
	\scalebox{1.0}{
		\begin{tabular}{llll}
			\hline
			& \textbf{Training}   & \textbf{Validation}  & \textbf{Testing}  \\ \hline
			$\#$ of sequences  & $300$   & $35$  & $70$ \\ %\hline
			$\#$ of time-variant sequences & $27$  & $3$  & $7$  \\ %\hline
			$\#$ of time-invariant sequences & $273$  & $32$  &  $63$ \\ %\hline
			Avg time span length (years) & $13.1$  & $16.0$  & $13.8$  \\ \hline
			% $\#$ of fundus images & $2646$   & $337$  & $688$ \\ %\hline
			% $\#$ of glaucoma fundus images & $110$  & $15$  & $28$  \\ %\hline
			% $\#$ of non-glaucoma fundus images & $2536$  & $322$  &  $660$ \\ \hline
			$\#$ of glaucoma fundus clips & $59$  & $9$  & $16$  \\ %\hline
			$\#$ of non-glaucoma fundus clips & $1087$  & $153$  &  $335$ \\ \hline
		\end{tabular}
	}
	\label{table:SIGF_stat}
\end{table}

\subsubsection{ADNI dataset}
To further validate generalization capability of the proposed method, we additionally leverage another longitudinal dataset with a completely different modality, i.e., \href{http://adni.loni.usc.edu}{Alzheimer's Disease Neuroimaging Initiative (ADNI)} database with 3D MRIs, and conduct comprehensive experiments and analysis. This dataset was launched to measure the progression of mild cognitive impairment (MCI) and early Alzheimer's disease (AD). 
Each subject is studied over a long-term period in this dataset. Specifically, the AD subjects are studied at $0$, $6$, $12$, and $24$ months, while MCI subjects who process a high risk for transition to AD are studied at $0$, $6$, $12$, $18$, $24$, $36$ months. In this work, we use MRI scans from ADNI1 ADNIGO and ADNI2 studies, and all the data underwent longitudinal post-processing operations using the FreeSurfer toolbox~\cite{reuter2012within}. We summarize the statistics of the ADNI dataset in Table~\ref{table:ADNI_stat}.
\begin{table}[!h]
	\centering
	\caption{Summary statistics of the ADNI dataset before and after data preprocessing. NC, MCI, AD are short for normal controls, mild cognitive impairment, and Alzheimer’s disease, respectively.} 
	\scalebox{1.0}{
		\begin{tabular}{llll}
			\hline
			& \textbf{Training}   & \textbf{Validation}  & \textbf{Testing}  \\ \hline
			$\#$ of subjects  & $496$   & $33$  & $133$ \\ %\hline
			$\#$ of time-variant subjects & $160$  & $11$  & $43$  \\ %\hline
			$\#$ of time-invariant subjects & $336$  & $22$  &  $90$ \\ %\hline
			Avg time span length (months) & $13.1$  & $14.0$ & $12.0$  \\ \hline
			% $\#$ of NC scans & $634$   & $36$  & $172$ \\ %\hline
			% $\#$ of MCI scans & $889$  & $73$  & $231$  \\ %\hline
			% $\#$ of AD scans & $576$  & $40$  &  $119$ \\ \hline
			$\#$ of NC clips & $504$   & $84$  & $137$ \\ %\hline
			$\#$ of MCI clips & $834$  & $139$  & $189$  \\ %\hline
			$\#$ of AD clips & $447$  & $68$  &  $64$ \\ \hline
		\end{tabular}
	}
	\label{table:ADNI_stat}
\end{table}

\subsubsection{Evaluation metrics}
To evaluate the performance of the proposed method, we utilize accuracy (ACC), sensitivity (SEN), specificity (SPE), and Area under the ROC Curve (AUC) for measurement. Note that the ADNI dataset contains three categories (NC, MCI, and AD), we calculate each evaluation metric value using the \textit{one-vs-one} approach and then take the macro-average.

\subsection{Implementation Details}
We implement the proposed MST-former with Pytorch and run experiments on a workstation with one NVIDIA GeForce RTX 4090 graphics card of 24 GB GPU memory and 2.10 GHz Intel Xeon Silver 4310 CPU. 
We equip the MST-former with $S=3$ scales
and let the scaling factor $\gamma=2$ for scale transition.
We set the temperature scalar $\tau$ in Eq.~\eqref{eq:balancedCE} as $2.0$ and the hyper-parameter controlling the time scaling attention in Eq.~\eqref{eq:t_matrix} are set as $(\alpha, \beta) = (0.5, 0.5)$.
As the two longitudinal datasets differs in many aspects, such as the sequence length and imaging modality, simply adoption of the same experimental settings would potentially result in sub-optimal performance. In this regard, we configure experiments on the two datasets separately to largely maximize of the model's potency.

\subsubsection{Experimental configurations on the SIGF dataset}
We adopt an overlapping approach to increase the sample size of the SIGF dataset~\cite{liu:20:deepGF}. Specifically, each complete sequence is segmented into equal-length sub-sequences (i.e., $6$) with a stride of $1$, resulting in $1146$, $162$, and $351$ clips for training, validation, and testing, respectively. It is worth noting that the class distribution is severely imbalanced (i.e., negative/positive=19:1), as shown in Table~\ref{table:SIGF_stat}.
We employ the Stochastic Gradient Descent (SGD) optimizer to update all learnable parameters with the momentum factor of $0.9$. The learning rate is initially set as $3\times 10^{-4}$ and varies in a linear warmup way followed by a cosine decay schedule. The training batch size is $4$. We run $300$ epochs to ensure model convergence. 

\subsubsection{Experimental configuration on the ADNI dataset}
Similar to the SIGF experiment, we convert each complete sequence into several equal-length clips. Each clip contains $3$ MRI scans with the size of $256\times 256 \times 256$. The detailed information of the preprocessed dataset can be found in Table~\ref{table:ADNI_stat}. The learning rate keeps $3\times 10^{-4}$ during the whole training period. We adopt the same optimizer as the SIGF experiment, and the training batch size is set as $2$.

\subsection{Experimental Results on the SIGF Dataset}
We compare the proposed method with $5$ state-of-the-art methods that are most relevant to this work, including AG-CNN~\cite{li2019attention}, Deep CNN~\cite{chen2015glaucoma}, DeepGF~\cite{liu:20:deepGF},
% transformer~\cite{vaswani2017attention}, 
Uniformer~\cite{li2201uniformer}, and GLIM-Net~\cite{xiaoyan:23:glim}. 
Among them, AG-CNN and Deep CNN are two networks designed for glaucoma detection. They take a single image as input and give the prediction of the next image. DeepGF involves the LSTM architecture for glaucoma forecasting. 
Uniformer and GLIM-Net are both transformer-based networks. Uniformer combines the merits of both convolution and self-attention and proposes dynamic position embedding (DPE) to encode 3D position information. It has exhibited outstanding performance across multiple tasks, such as video classification, dense prediction, etc. GLIM-Net is specifically designed for glaucoma forecasting, which recently achieves state-of-the-art results on the SIGF dataset.

\begin{figure}[htbp]
	% \begin{strip}
		\centering
		\includegraphics[width=0.4\textwidth]{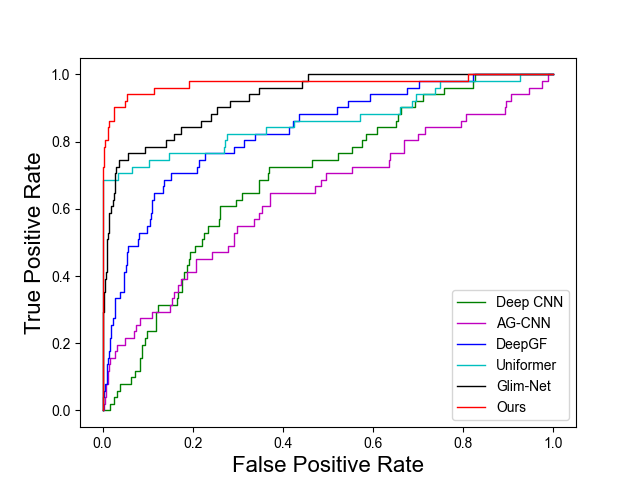}	
		\caption{ROC curves of Deep CNN, AG-CNN, Naive transformer, DeepGF, GLIM-Net, Uniformer, and MST-former (ours) on the SIGF dataset.}
		\label{fig:ROC_SIGF}  
		% \end{strip}
\end{figure}

We summarized the numerical results of the comparison experiment in Table~\ref{table:forecast_results_SIGF}. Deep CNN and AG-CNN have the least satisfactory results, with AUC of 0.696 and 0.640 respectively, indicating that the information of a single image is far insufficient to make reliable disease forecasting.
DeepGF largely improves the performance (AUC of 0.850) by employing LSTM to integrate features of the sequential data, which verifies the importance of the temporal information. All the transformer-based approaches, including Uniformer, GLIM-Net, and the proposed MST-former have superior performance to DeepGF, which implies that the transformer structure is better at longitudinal learning than LSTM. 
Although Uniformer combines the convolution to reinforce the spatial representation and incorporates temporal information into the DPE module, it does not address the issue of irregular sampling. Therefore, its performance is impaired when it is generalized to the irregular time series data dataset. 
GLIM-Net proposes time-sensitive time-sensitive multi-head self-attention modules to address the irregularly sampled data, which largely boosts the forecasting performance, achieving AUC of 0.931. However, GLIM-Net crudely translates each whole image into a feature vector, the representation ability of the image features might be limited and insufficient.
Noticeably, the proposed MST-former achieves the best performance, with AUC of 0.986, accuracy of 97.1\%, sensitivity of 94.1\%, and specificity of 97.2\%, outperforming compared methods substantially. Specifically, compared with the latest approach GLIM-Net, our proposed method improves AUC, accuracy, sensitivity, and specificity by $5.5\%$, $10.4\%$, $7.6\%$, $9.2\%$ in absolute terms, respectively. 
Besides, the proposed network has a much fewer number of parameters (26.2M) than GLIM-Net does (45.0M).
The ROC curves of different methods are shown in Fig.~\ref{fig:ROC_SIGF}.

\begin{table}[!h]
	\centering
	\caption{Comparison results on the SIGF dataset in terms of AUC, ACC (Accuracy), SEN (Sensitivity), and SPE (Specificity) criterion.We also list the number of trainable parameters in each model in millions (M).}
	\setlength{\tabcolsep}{3pt}
	\scalebox{1.0}{
		\begin{tabular}{lccccc}
			\hline
			Method & \textbf{AUC} & \textbf{ACC ($\%$)}   & \textbf{SEN ($\%$)}  & \textbf{SPE ($\%$)}   & Params \\ \hline
			Deep CNN~\cite{chen2015glaucoma} &  $0.696$ & $69.7$  &  $62.7$   &  $69.9$ & $1.9$M  \\ %\hline
			% transformer~\cite{vaswani2017attention} & $0.866$ & $83.3$   & $76.5$   & $83.5$    \\
			AG-CNN~\cite{li2019attention} &  $0.640$ & $55.7$ & $64.7$  & $55.4$ & $3.0$M \\ %\hline
			DeepGF~\cite{liu:20:deepGF} & $0.850$ & $76.0$  & $79.4$   & $75.9$  & $4.8$M \\ %\hline
			Uniformer~\cite{li2201uniformer} & $0.863$ & $95.5$   & $70.6$   & $96.2$  & $22.0$M \\
			GLIM-Net~\cite{xiaoyan:23:glim} & $0.931$ & $86.7$  & $86.5$   & $88.0$ & $45.0$M \\  \hline
			\textbf{Ours} & $\textbf{0.986}$ & $\textbf{97.1}$  & $\textbf{94.1}$  &  $\textbf{97.2}$ & $26.2$M\\ \hline % 282
		\end{tabular}
	}
	\label{table:forecast_results_SIGF}
\end{table}

\subsection{Experimental Results on the ADNI dataset}
\begin{figure*}[htbp]
	\centering
	\subfigure[GLIM-Net]{\includegraphics[width=0.32\textwidth]{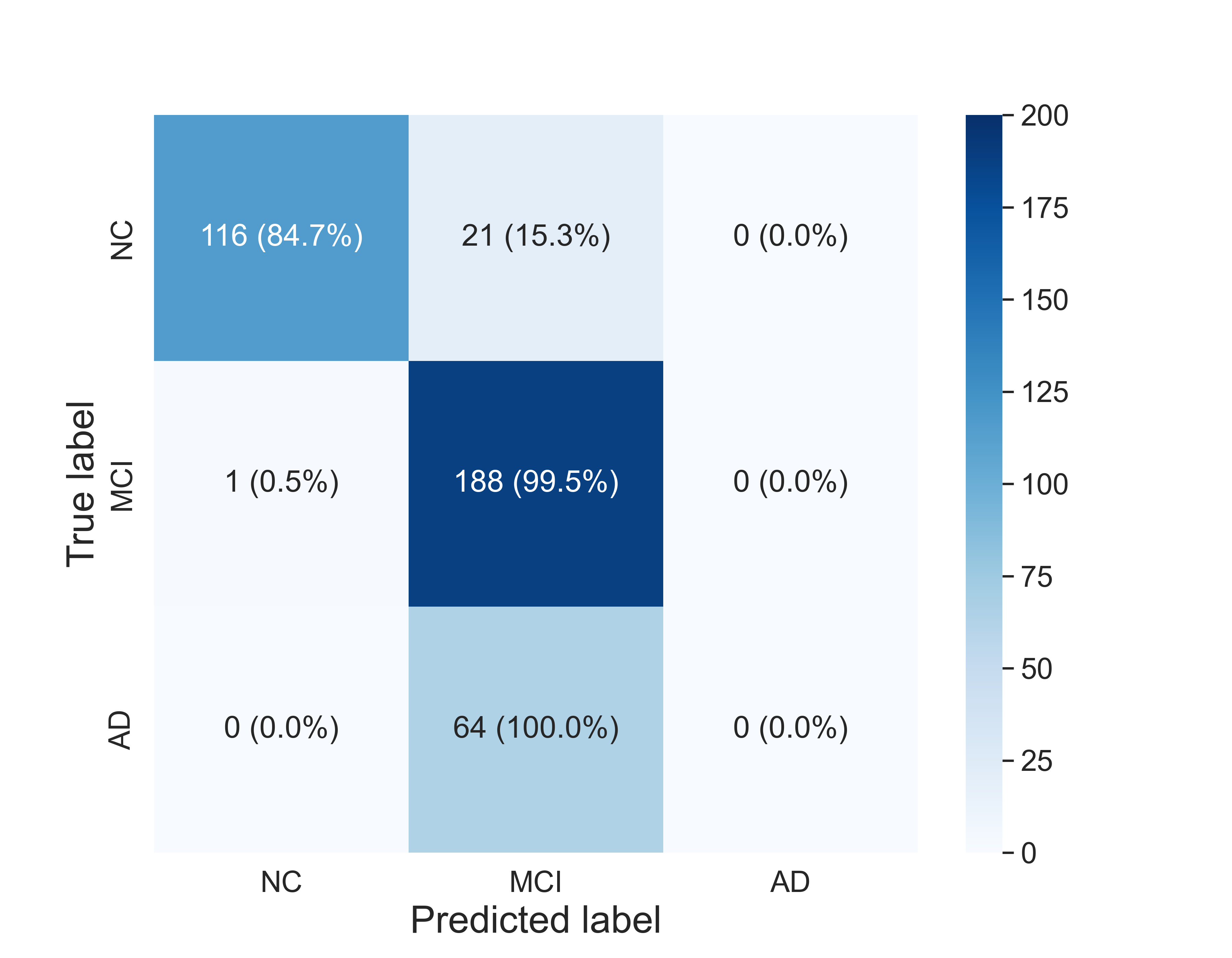}} 
	\subfigure[Uniformer]{\includegraphics[width=0.32\textwidth]{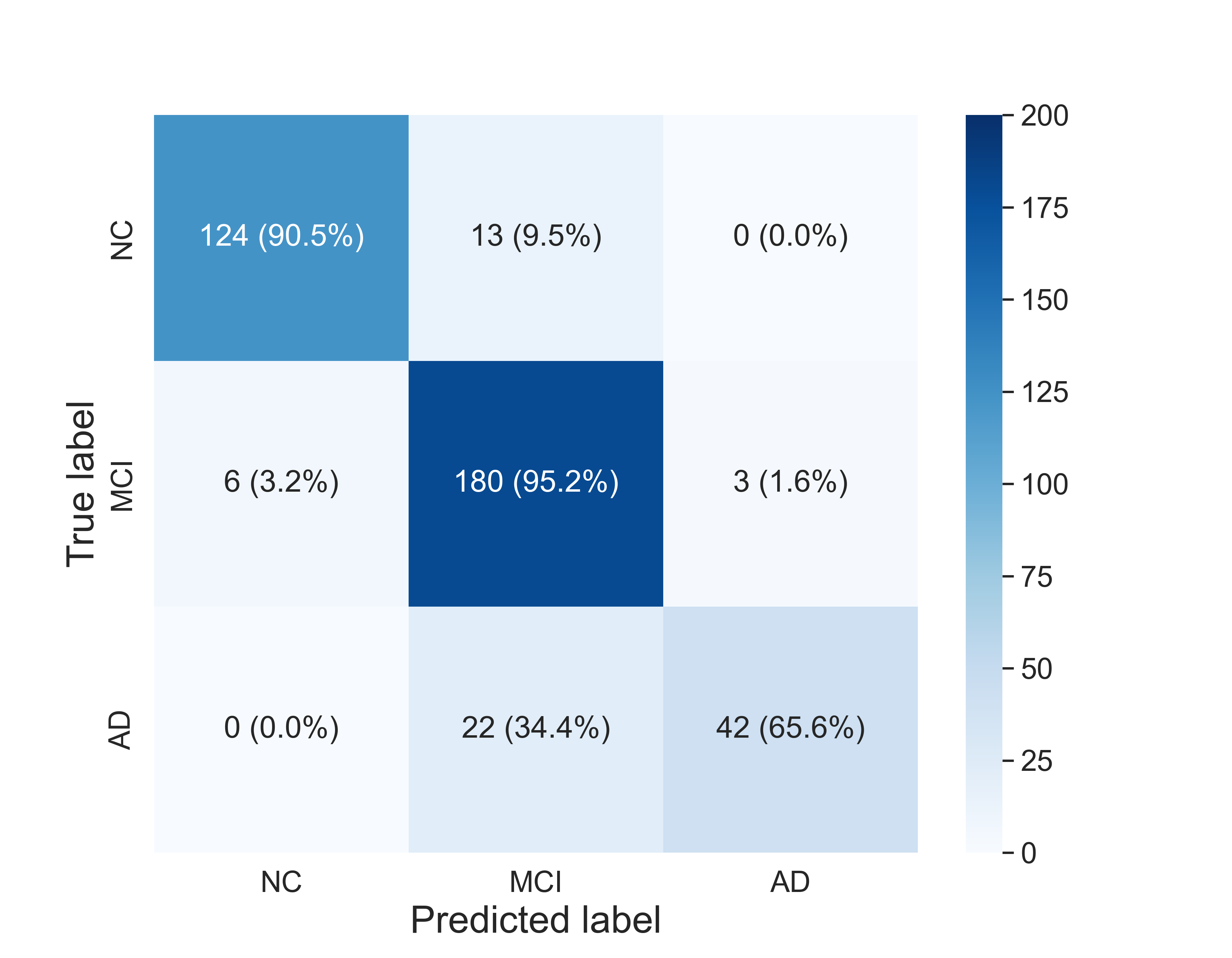}} 
	\subfigure[MST-former]{\includegraphics[width=0.32\textwidth]{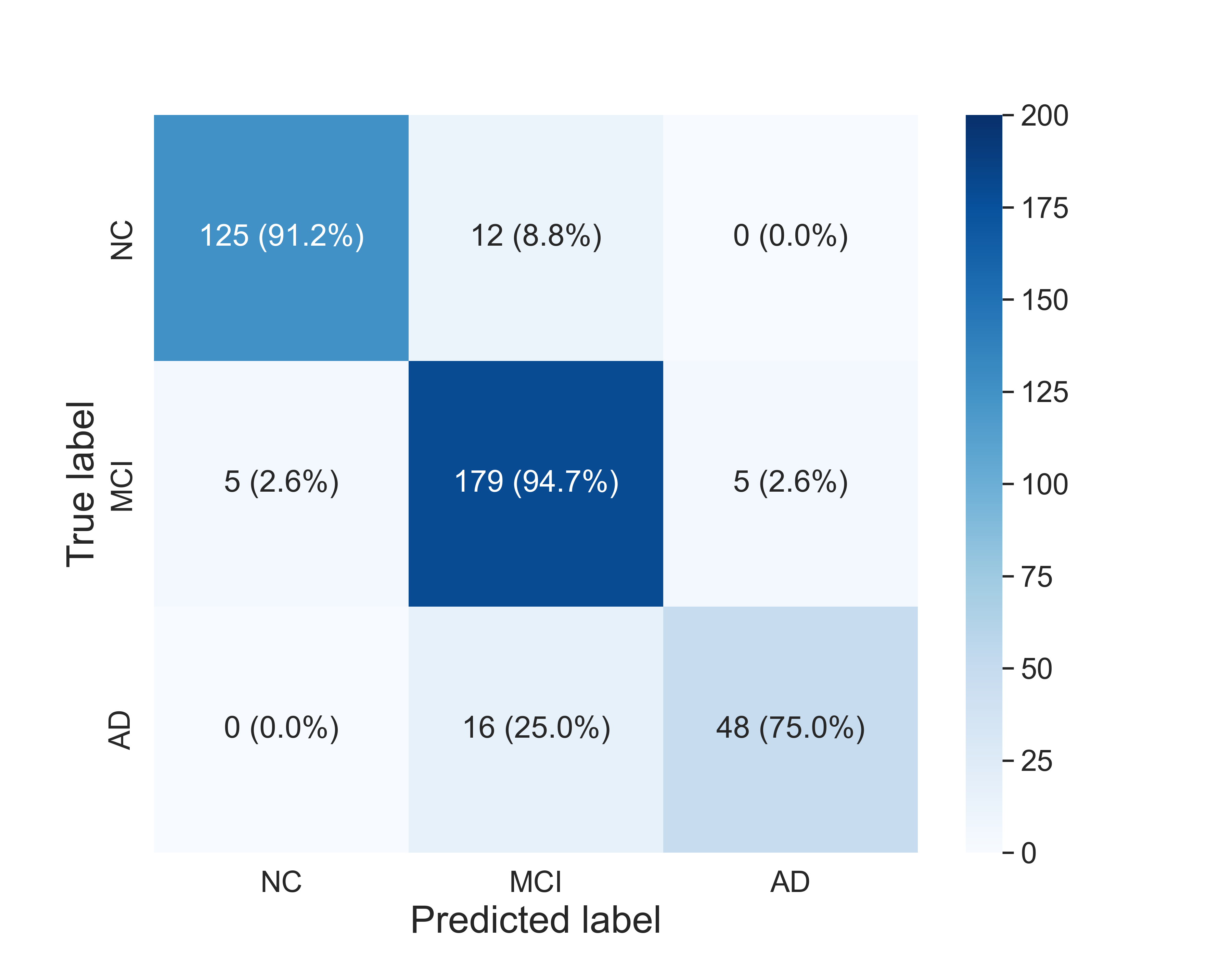}}
	\caption{Confusion matrix of GLIM-Net (left), Uniformer (middle), MST-former (right) for ADNI experiment. The integer values in the matrix represent the number of samples belonging to different classes with the corresponding percentage indicating the ratio in the predicted label.}
	\label{fig:adni_con_matrix}
\end{figure*}

Next, we investigate the performance of the MST-former on the ADNI dataset. Similarly, given a sequence of image frames composed of MRI scans recorded at different time points, our goal is to predict the subject's next status as either NC, MCI, or AD. Based upon the experimental results on the SIGF datasets, we select the top two approaches, Uniformer and GLIM-Net for comparison. As shown in Table~\ref{table:results_ADNI}, our approach consistently exhibits great advantage over GLIM-net and Uniformer, with an improvement of 10.0\% and 5.9\% on AUC, respectively.  
Besides, we compute the confusion matrix of GLIM-net, Uniformer, and MST-former, which is shown in Fig.~\ref{fig:adni_con_matrix}. From Table~\ref{table:results_ADNI}, we observe that there is a tremendous performance degradation on GLIM-Net, with a poor macro AUC of 0.866, accuracy of 77.9\%, sensitivity of 61.4\% and specificity of 85.8\%. In comparison to the 2D fundus images, the brain MRI scans have relatively much larger size. The pathogenesis of cognitive impairment and Alzheimer's disease is complicated, and the symptoms might sparsely distribute over the whole 3D MRI scan. Simply leveraging a CNN to convert the whole MRI scan into a feature vector by GLIM-Net cannot adequately exploit the most discriminative regions for feature representation, which results in inferior performance, especially on AD class that has the least samples. In contrast, the proposed MST-former exhibits stable and outstanding performance due to its strong capability to capture representative multi-scale intra-image features spatially and inter-image temporally, which outperforms Uniformer considerably on all evaluation metrics.
It is worth noting that our method has a much higher sensitivity in forecasting the most difficult category, i.e., 75\% on AD, compared to 0\% and 65.6\% obtained by GLIM-Net and Uniformer (Fig.~\ref{fig:adni_con_matrix}). This is of great clinical significance in early detection potential AD patients, and promoting precautions to prevent further deterioration.

\begin{table}[!h]
	\centering
	\caption{Comparison results on the ADNI dataset in terms of AUC, ACC (Accuracy), SEN (Sensitivity), and SPE (Specificity) criterion.}
	% recall = sensitivity
	\setlength{\tabcolsep}{3pt}
	\scalebox{1.0}{
		\begin{tabular}{lccccc}
			\hline
			Method & \textbf{AUC}  & \textbf{ACC ($\%$)}   & \textbf{SEN ($\%$)}  & \textbf{SPE ($\%$)} \\ \hline
			Uniformer~\cite{li2201uniformer} & $0.907$ & $88.7$  & $83.8$   & $93.1$  \\  % 22.0M 
			GLIM-Net~\cite{xiaoyan:23:glim} & $0.866$ & $77.9$  & $61.4$   & $85.8$ \\  % 45.8M
			\hline
			\textbf{Ours} & $\textbf{0.966}$  & $\textbf{90.3}$  & $\textbf{87.0}$  & $\textbf{94.2}$   \\ \hline % 46.8M 
		\end{tabular}
	}
	\label{table:results_ADNI}
\end{table}

\subsection{Ablation Study}
To verify the rationality of our design, we conduct the ablation study on the three proposed components, the space-time positional encoding (STP), time-aware temporal attention block (TTA), and multi-scale setting (MS) on the SIGF dataset. Note that the removal of multi-scale setting means that we do not perform the scale transition operation and set the number of encoder-decoder blocks to 3. 
The experimental results are reported in Table~\ref{table:ablation_SIGF}.
It is noticeably that the model's performance is least satisfactory when all the three components are missing (model \ding{172}), with AUC of 0.913 and accuracy of 81.1\% only. 
When solely equipped with any one of the three components (models \ding{173}, \ding{174}, and \ding{175}), the model can improve the AUC by 2.5-3.4\% in absolute terms. It is worth mentioning that the three models have already outperformed the current state-of-the-art method GLIN-Net (AUC of 0.931).
Besides, when one more other component is added, the model's performance is further boosted (models \ding{176}, \ding{177}, and \ding{178}).
With STP, TTA and MS colaborate together, the model improves all the evaluation metrics to reaching the highest values.

\begin{table}[!h]
	\centering
	\caption{Ablation on space-time positional encoding (STP), time-aware temporal attention (TTA), and muti-scale setting (MS).}
	\scalebox{1.0}{
		\begin{tabular}{c|c|c|c|cccc}
			\hline
			&STP & TTA & MS  & \textbf{AUC}  & \textbf{ACC ($\%$)}   & \textbf{SEN ($\%$)}  & \textbf{SPE ($\%$)}  \\ \hline
			\ding{172}&\ding{55} & \ding{55} & \ding{55} & $0.913$  & $81.1$  & $86.3$ & $80.9$ \\ % 242
			\ding{173}&\ding{55} & \ding{55} & \ding{51}  & $0.938$  & $82.0$  & $88.2$ & $81.8$ \\
			\ding{174}&\ding{55} & \ding{51} & \ding{55} & $0.943$ & $82.0$ & $90.2$  & $81.7$\\  % 161
			\ding{175}&\ding{51} & \ding{55} & \ding{55} & $0.945$ & $83.1$ & $84.3$  & $82.7$\\  %245
			\ding{176}&\ding{51} & \ding{51} & \ding{55} & $0.971$ & $96.5$ & $88.2$  & $96.8$\\  % 171
			\ding{177}&\ding{55} & \ding{51} & \ding{51} & $0.962$ & $91.8$   & $86.3$  & $92.0$  \\ %\hline % 89
			\ding{178}&\ding{51} & \ding{55} & \ding{51} & $0.961$   & $87.9$  & $92.2$   &  $87.8$   \\  \hline
			\ding{179}&\ding{51} & \ding{51} & \ding{51} & $\textbf{0.986}$ & $\textbf{97.1}$  & $\textbf{94.1}$  & $\textbf{97.2}$  \\ \hline
		\end{tabular}
	}
	\label{table:ablation_SIGF}
\end{table}

\subsection{Varying the Number of Scales}
The multi-scale setting enables our model to capture features at different spatial resolutions, Here, we investigate the influence of the number of scales on the performance of glaucoma forecasting. Note that each scale includes just one encoder-decoder block. Table~\ref{table:vary_layer_SIGF} reports the results of MST-former equipped with different numbers of scales on the SIGF dataset. 
We observe that with the number of scales increasing from 1 to 3, the model's performance improved steadily, with AUC from 0.648 to 0.986 reaching the summit. However, when we further add more scales (i.e., 4), there is a slight performance drop. On the other hand, involving more scales will inevitably introduce more learnable parameters, causing more computational burden. As MST-former with 3 scales achieves the best performance with a moderate number of parameters, we choose such setting for the majority of our experiments. 

\begin{table}[!h]
	\centering
	\caption{AUC, Accuracy (ACC), sensitivity (SEN), and specificity (SPE) on SIGF dataset with different numbers of scales. The number of parameters is also presented.} 
	\scalebox{1.0}{
		\begin{tabular}{lcccccc}
			\hline
			& \textbf{AUC} & \textbf{ACC ($\%$)}   & \textbf{SEN ($\%$)}  & \textbf{SPE ($\%$)}    & Params\\ \hline
			$1$ Scale & $0.648$ & $60.6$  & $60.8$  & $61.4$     & $10.4$M \\ %\hline
			$2$ Scales & $0.828$  & $79.8$  &  $82.4$ &  $79.2$  & $18.3$M\\ %\hline  % 167 0.0003
			$3$ Scales & $\textbf{0.986}$ & $\textbf{97.1}$  & $\textbf{94.1}$  &  $\textbf{97.2}$  & $26.2$M \\ %\hline
			$4$ Scales & $0.981$  & $96.6$  & $\textbf{94.1}$  & $96.6$  & $34.1$M \\ \hline % 153 0.0001
		\end{tabular}
	}
	\label{table:vary_layer_SIGF}
\end{table}

\subsection{Efficacy of Balanced Softmax Cross-entropy Loss}
Lastly, we inspect the efficacy of the proposed temperature controlled Balanced Softmax Cross-entropy loss on alleviating the class imbalanced problem, and how the value of the temperature parameter $\tau$ affects the model's performance.
We vary the value of the temperature parameter $\tau$ in the range of $\{1.00, 1.25, 1.50, 1.75, 2.00, 2.25, 2.50\}$, and the corresponding results are displayed in Table~\ref{table:loss_change}.
In comparison to the cross-entropy loss, the standard Balanced Softmax Cross-entropy loss ($\tau=1$) shows far better predictive ability, demonstrating over 10\% performance gain in all evaluation metrics. When the temperature controller is introduced (i.e., $\tau>1$), the results are generally improved significantly. 
As $\tau$ gradually increases from 1.00 to 2.00, Balanced Softmax Cross-entropy loss can gradually improve the model's performance by alleviating differences in label distribution, with AUC gradually rising and peaking at 0.986. However, the advantage begins to weaken when $\tau$ surpasses 2.00 (e.g., 2.25 and 2.50). Hence, we fix $\tau=2.00$ in this work.

\begin{table}[!h]
	\centering
	\caption{Comparison of cross-entropy (CE) loss and Balanced Softmax Cross-entropy (Balanced Softmax CE) loss with different temperature values. The evaluation is conducted under the SIGF dataset.} 
	\scalebox{0.9}{
		\begin{tabular}{l|l|ccccc}
			\hline
			\multicolumn{2}{c|}{\textbf{Loss}} & \textbf{AUC ($\%$)} & \textbf{ACC ($\%$)}   & \textbf{SEN ($\%$)}  & \textbf{SPE ($\%$)}    \\ \hline
			\multicolumn{2}{c|}{CE}  & $0.842$  & $83.1$  & $74.5$  & $83.4$    \\ \hline
			\multirow{7}*{\makecell[c]{Balanced \\ Softmax CE}} &$\tau=1.00$ & $0.945$ & $95.8$  & $78.4$  & $96.3$ \\
			~ &$\tau=1.25$ &$0.952$  & $97.2$  & $80.3$  & $92.5$ \\
			% ~ &$\tau=1.75$  225 \\ 
			~ &$\tau=1.50$ & $0.957$  & $98.8$  & $76.5$  & $99.0$     \\ % 232
			~ &$\tau=1.75$ & $0.972$  & $98.3$ & $78.4$ & $98.6$ \\ % 226
			~ &$\tau=2.00$ & $\textbf{0.986}$ & $97.1$  & $94.1$  &  $97.2$  \\
			~ &$\tau=2.25$  & $0.983$  & $83.9$  & $96.1$  & $83.6$   \\ % 268
			~ &$\tau=2.50$  & $0.963$  & $92.9$  & $90.2$  & $93.1$   \\ \hline 
		\end{tabular}
	}
	\label{table:loss_change}  
\end{table}

\section{Discussion}\label{sec:discussion}

Making effective predictions for glaucoma before the onset of the disease has significant benefits for patients. In this work, we introduce the Multi-scale Spatio-temporal Transformer Network (MST-former) to forecast the probability of future glaucoma occurrence based on a series of historical fundus images. 
Rather than GLIM-Net \cite{xiaoyan:23:glim} that isolated spatial feature extraction from temporal modeling, potentially restricting the performance of their model, our MST-former uses the multi-head spatial-temporal attention module stacked in multiple scales to concurrently capture intra-image features spatially and inter-image variation temporally. The proposed space-time positional encoding leverages the geometric information of the sequential image input in the spatio-temporal dimension. We first evaluate the proposed method on the SIGF dataset for glaucoma forecasting, which exhibits superior performance compared to other advanced methods. Then we apply it to another task using a dataset of a completely distinct imaging modality, i.e., ADNI dataset, for mild recognition impairment and Alzheimer’s Disease prediction. Consistently, the proposed method also achieves outstanding performance, indicating its excellent generalization ability. 

Besides, the three key components, i.e., space-time positional encoding (STP) module, time-aware temporal self-attention and the multi-scale encoder-decoder architecture, play important roles in enhancing the model's discrimination ability, which is verified by our ablation study. As expected, the proposed method gains the maximum benefit if and only if the three components cooperate together. 
In addition, we investigate how the multi-scale setting of the encoder-decoder block influences the model performance by varying the number of scales through grid search.
We observe that increasing the number of scales does not necessarily guarantee an improvement of model performance. For instance, when the number exceeds 3, the model' performance slightly degrades.
Such performance saturation phenomenon might be attributed to the overfitting issue. Given limited training samples, over-complex network (e.g. MST-former with 4-scale encoder-decoder blocks) might easily capture noise instead of the underlying signal.
Furthermore, we also inspect the optimal temperature regulator $\tau$ in the Balanced Softmax Cross-entropy loss by grid search, and prove that imposing the temperature regulator indeed boosts the model performance, showing prominent advantage over the vanilla version (i.e., $\tau=1$).

Despite its prominent performance, the proposed method has some limitations. First, the current version of the proposed model could just forecast the probability of developing glaucoma for an unspecified time. However, it is more significant to predict its likelihood conditioned on a specific time to facilitate timely precautions against further vision loss or even blindness. One possible solution is to impose a time condition in both the encoder and the decoder
as additional input. Second, the scaling factor $\gamma$ across all the scale transition processes is fixed as 2 in this study. Whether selection of other values or dynamically changing $\gamma$ is not verified at present, which is worthy of trying in future work.  
Third, the conventional evaluation metrics for disease forecasting might be sub-optimal. Note that there is skewed data distribution in most of the longitudinal medical datasets, i.e., the time-invariant sequential data severely overwhelms the time-variant ones, thus resulting in significant bias in model evaluation. How to effectively assess a model's predictive capability under a dataset with few time-variant imaging inspections is important to be further excavated.
Forth, the proposed method is particularly designed for the sequential medical images of large size, and might not be well adopted to those of relatively small size (e.g., cell images). In comparison to using a standard convolutional stem for extracting spatial information, the advantages of computing spatial attention on small-size images, e.g. $32\times 32$ are not very pronounced.

In the future, we will continue exploring the potential of the proposed method for modeling irregular time-series images and extend it for multi-modal learning. For instance, it is promising to leverage multiple modalities like MRI, PET, other biological markers, and clinical and neuropsychological assessment of the ADNI dataset to boost the forecasting performance.
Additionally, we aspire to collect and find more similar datasets to validate the effectiveness of our approach. 
We also plan to develop a fully automated pipeline including longitudinal pre-processing, time-series representation, and disease forecasting for various diseases.

\section{Conclusion}\label{sec:conclusion}

In this work, we present a innovative class imbalanced longitudinal learning framework, multi-scale spatio-temporal transformer network, for glaucoma forecasting, which can handle irregular time series images with imbalanced class distribution. The proposed spatio-temporal self-attention can simultaneously exploit intra-image and inter-image semantic information from longitudinal images, and the multi-scale setting can facilitate learning at various spatial resolutions. To handle input with irregularly sampled images, we incorporate time-aware temporal attention, weighting temporal attention with the time distance matrix. Moreover, we devise the temperature-controlled Balanced Softmax Cross-entropy loss to solve the imbalanced issue. 
Extensive experiments on the SIGF dataset for glaucoma forecasting demonstrate 
that the proposed method outperforms other methods substantially in terms of all evaluation metrics. The efficacy of the key components is also verified by a series of alabtion studies. Furthermore, our method achieves state-of-the-art results on the ADNI MRI dataset for prediction of mild recognition impairment and Alzheimer's disease, suggesting its strong stability and robustness. We hope that this work can provide researchers with new insight into disease forecasting and promote more effective approaches and clinical applications in this direction.

\section*{References}
\bibliographystyle{IEEEbib}	
\footnotesize
\small
\bibliography{refs}

\end{document}